\def\year{2018}\relax
\documentclass[letterpaper]{article} 
\usepackage{aaai18}  
\usepackage{times}  
\usepackage{helvet}  
\usepackage{courier}  
\usepackage{url}  
\usepackage{graphicx}  

\usepackage{amsmath,amssymb,amsthm}
\usepackage{adjustbox}
\usepackage{amsfonts}
\usepackage{booktabs}
\usepackage[ruled,vlined]{algorithm2e}
\usepackage{svg}
\usepackage{tikz}
\usepackage[english]{babel}
\usepackage[utf8]{inputenc}
\usepackage[noend]{algpseudocode}
\definecolor{green}{RGB}{0,95,0}
\definecolor{red}{RGB}{148,150,152}
\DeclareMathOperator*{\argmax}{arg\,max}
\usepackage{multirow}
\usepackage{array}
\usepackage{multirow}
\usepackage{booktabs}

\newcommand\norm[1]{\left\lVert#1\right\rVert}
\newcommand{\model}{FastKATE}

\usetikzlibrary{shapes}
\usetikzlibrary{fit}
\usetikzlibrary{calc}
\tikzset{
myrect/.style={draw=black, rectangle},
myarrow/.style={-latex}
}

\title{Fast Top-$\boldsymbol{k}$ Area Topics Extraction with Knowledge Base}
\author{Fang Zhang$^\dagger$, Xiaochen Wang$^\dagger$, Jingfei Han$^\ddagger$, Jie Tang$^\dagger$, Shiyin Wang$^\dagger$, Marie-Francine Moens$^\sharp$\\
$^\dagger$Department of Computer Science and Technology, Tsinghua University\\
$^\ddagger$Department of Computer Science and Technology, Beijing University of Aeronautics and Astronautics\\
$^\sharp$Department of Computer Science, KU Leuven\\
\{fang-zha15, xiaochen15, wangshiy16\}@mails.tsinghua.edu.cn, jfhan@buaa.edu.cn, jietang@tsighua.edu.cn, sien.moens@cs.kuleuven.be\\
}

\begin{document}
%
\maketitle
\begin{abstract}
What are the most popular research topics in Artificial Intelligence (AI)?
We formulate the problem as extracting top-$k$ topics that can best represent a given area with the help of knowledge base. We theoretically prove that the problem is NP-hard and propose an optimization model, \model, to address this problem by combining both explicit and latent representations for each topic. We leverage a large-scale knowledge base (Wikipedia) to generate topic embeddings using neural networks and use this kind of representations to help capture the representativeness of topics for given areas. We develop a fast heuristic algorithm to efficiently solve the problem with a provable error bound. We evaluate the proposed model on three real-world datasets. Experimental results demonstrate our model's effectiveness, robustness, real-timeness (return results in $<1$s), and its superiority over several alternative methods.

\end{abstract}

\section{Introduction}
Automatically extracting top-$k$ topics of a given area is fundamental in the historical analysis of the given area. With the ability of solving this problem, not only can we gain an accurate overview of the given area, but it can also help make our society more efficient, such as giving suggestions on how to optimize the allocation of resources (e.g., research fundings) to more representative and important topics. This can also provide guidances to newcomers of the area. However, there are too many topics in almost any areas, and for any researcher, it is non-trivial for him/her to extract the top-$k$ topics of the given area in a short period of time, especially if the researcher is a newcomer to the area. Therefore it is important to find a way to automatically solve this problem.

While much research has been conducted on the topic extraction problem, their main focus is basically on \textbf{document} topic extraction, but not on \textbf{area} topic extraction. For example, in ~\cite{blei2003latent,griffiths2004finding}, latent dirichlet allocation (LDA) model is used to model topics in documents and abstracts, where topics are represented as multinomial distributions over words. Topics can also be represented as keyphrases (or topical phrases), and under this perspective, keyphrases extraction task can also be viewed as topic extraction task. Different models such as frequency-based~\cite{Salton1997Term}, graph-based~\cite{mihalcea2004textrank}, clustering-based~\cite{grineva2009extracting} and so on have been explored to address the keyphrase extraction problem, but still focus on a document instead of an area.

The problem of area topic extraction is novel, non-trivial and poses a set of unique challenges as follows: (1) How to formulate the problem and using what kind of datasets and how to use it is not clear. (2) How to capture the representativeness of topics for a given area is another challenging issue. (3) The number of candidate topics in a given area may be very large. There are 14,449,404 page titles (including categories) in Wikipedia and even after we do some preprocessing on it, we still get 9,355,550 topics. Thus how to develop an efficient algorithm to apply it in practice is important too. (4) Since there are no standard benchmarks that can perfectly match this problem, how to quantitatively evaluate the results is also a challenging issue.

To address these challenges, in this paper, we give a formal definition of the problem and develop an optimization model to efficiently solve it. Our contributions can be summarized as follows:
\begin{itemize}
  \item To the best of our knowledge, this is the first attempt to formulate and address the area topic extraction problem. We formulate the problem as extracting top-$k$ topics that can best represent a given area with the help of knowledge base. We theoretically prove that the problem is NP-hard.
  \item We propose an optimization model, \model, to address this problem by combining both explicit and latent representations for each topic. We leverage a large-scale knowledge base (Wikipedia) to generate topic embeddings using neural networks and use this kind of representations to help capture the representativeness of topics for given areas. We develop a fast heuristic algorithm to efficiently solve the problem with a provable error bound.
  \item We evaluate the proposed model on three real-world datasets. Experimental results demonstrate our model’s effectiveness, robustness, real-timeness (return results in $<1$s), and its superiority over several alternative methods.
\end{itemize}

\section{Problem Formulation} \label{sec:pf}
We first provide necessary definitions and then formally define the problem.

\paragraph{Definition 1.} \textbf{Knowledge Base and Topic.}
A knowledge base is represented as a triple $\mathit{KB} = (C, R, X)$, where $C$ represents a set of knowledge concepts, and we also view this as \textbf{topics} in this paper. $R$ represents a set of relations between topics. $X$ represents a set of co-existences between topics, i.e., each $x_i \in X$ is a sequence of topics $\{t_{i1}, t_{i2}, t_{i3}, ...\}$, where $t_{ij} \in C$.

This definition is a variation of that in~\cite{mcguinness2004owl,tang2015sampling}. In our work, $X$ represents a corpus consisting of massive documents, and each document is a sequence of topics. Relations $R$ may have various types; we focus on sub-topic and super-topic relations in our work.

Each topic $t \in C$ in a knowledge base $\mathit{KB}$ already has a corresponding topical phrase, such as ``Artificial Intelligence''. To help grasp the relations/similarities between these topical phrases, we also represent each topic $t \in C$ as a vector $\mathbf{v}_t \in \mathbb{R}^n$ in a latent feature space, where $n$ is the dimension of the feature space, which will be detailed in section~\ref{subsec:topic_embedding}. Thus each topic in our work has both explicit representation (i.e., topical phrase) and latent representation (i.e., vector).

\paragraph{Definition 2.} \textbf{Area.}
In this paper, an area $r$ is essentially also a topic in $C$. Thus it has the same form and attributes as other topics in $C$. An area may be also a topic of some other area. For example, Machine Learning is an area, and it can also be viewed as a topic of Artificial Intelligence area.

We leverage a knowledge base to help extract topics from a given area in our work. We formally define the problem as follows.

\paragraph{Problem 1.} \textbf{Extracting top-$\boldsymbol{k}$ topics in a given area.}

The input of this problem includes an external knowledge base $\mathit{KB}=(C, R, X)$, a given area $r \in C$ and the number $k \in \mathbb{Z}^+$ of topics needed to be extracted.

The output of this problem is a set of top-$k$ topics $T^k=\{t \mid t \in C\}$ which can represent the given area best.

Our goal is to learn a function $f$ from the given input so as to extract the top-$k$ topics which can represent the given area best. More specifically, $f$ is defined as: 
\begin{equation}
\centering
\begin{split}
f: \big\{r, k, \mathit{KB}=(C,R,X) \mid r \in C, k \in \mathbb{Z}^+\big\} \\
 \mapsto \big\{T^k=\{t \mid t \in C\} \big\}.
\end{split}
\end{equation}

This problem is equivalent to selecting $k$ topics from the topics set $C$ that can represent the given area $r$ best. We use $D(T^k, C, r)$ to denote the degree of how well a set of $k$ topics $T^k \subseteq C$ can represent all topics in $C$ \textbf{on the given area $\boldsymbol{r}$}.
Without loss of generality, we assume $D(T^k, C, r) \ge 0$. And since adding new topics to $T^k$ should not reduce the representativeness of previous extracted topics, $D(T^k, C, r)$ should be monotonically non-decreasing. We also assume reasonably that topics added in early steps should not help (actually it may damage) topics added in later steps increase the value of the goal function. This means that a topic added in later steps contribute equal or possibly less to the goal function compared with that when the same topic is added in early steps. This is intuitive and reasonable because if a topic is added in later steps, some previous added topics may already have a good representation of the area, and thus this topic's contribution to the goal function may be decreased. We will show that this attribute implies the goal function's submodulariry~\cite{svitkina2011submodular} in the following section.
Then our problem can be reformulated as follows: $$T^{k*}={\argmax_{T^k \in C,|T^k|=k}}{D(T^k, C, r)},$$ where $D(T^k, C, r)$ is a non-negative and monotonically non-decreasing function.



\section{The Proposed Model} \label{sec:tpm}

We propose \textbf{\model} (\textbf{Fast} top-\textbf{K} \textbf{A}rea \textbf{T}opics \textbf{E}xtraction) to address the problem. In general, \model not only represents topics in explicit forms (phrases) as in knowledge bases, but also represents topics as vectors in a latent feature space, and uses a neural network-based method to learn topic embeddings from an external large-scale knowledge base. \model further incorporates domain knowledge from the knowledge base to assign ``general weights'' to different topics to help solve the problem. We develop a heuristic algorithm to efficiently solve the defined problem and we prove our algorithm is at least $(1-1/e)$ of the optimal solution. We further develop a fast implementation of our algorithm which can return results in real-time.


\subsection{Topics Representation} \label{subsec:topic_embedding}
We first generate $C$ and use it as candidate topics and then train embeddings $\mathbf{v}_t \in \mathbb{R}^n$ for each topic $t_i \in C$. We use Wikipedia as our knowledge base $\mathit{KB}$ to help generate candidate topics and train topics embeddings. We extract 14,449,404 titles of all articles and categories from Wikipedia, and convert them into lower forms and remove possible duplicates and those consisting of punctuations from these titles. Finally we get 9,355,550 titles as candidate topics. Then we use an unsupervised neural network-based method to learn the embeddings of these topics. We then preprocess the Wikipedia corpus to keep only candidate topics in the corpus, and use the preprocessed Wikipedia corpus as our training data. We adopt a similar method to that used in Word2Vec~\cite{mikolov2013distributed}. We treat each topic as a single token, and use a Skip-Gram model to generate each topic's embedding. In the Skip-Gram model, the training objective is to find topic embeddings that are useful for predicting surrounding topics. More formally, given a sequence of training topics, $t_1, t_2, t_3, ..., t_N$, the objective of the Skip-Gram model is to maximize the average log probability $$\frac{1}{N} \sum\limits_{i=1}^{N} \sum\limits_{-c \leq j \leq c, j \neq 0} \log p(t_{i+j}|t_i)$$ where $c$ is the size of the training context (also denoted as window size), and $p(t_{i+j}|t_i)$ is defined using the softmax function: $$p(t_O|t_I) = \frac{ \exp(\mathbf{v}_{t_O}^T \mathbf{v}_{t_I}) }{\sum\limits_{i=1}^{T} \exp(\mathbf{v}_i^T \mathbf{v}_{t_I})}$$ where $\mathbf{v}_{t_I}$ and $\mathbf{v}_{t_O}$ are the embeddings of ``input topic'' $t_I$ and ``output topic'' $t_O$ respectively, and $T$ is the number of total candidate topics. Because $T$ is very large, this calculation is very computationally expensive. Thus we adopt a common approximation in our model: Negative Sampling (NEG)~\cite{mikolov2013distributed}, which can speed up the training process greatly. Using NEG, $\log p(t_O|t_I)$ is replaced by: $$\log \sigma (\mathbf{v}_{t_O}^T \mathbf{v}_{t_I}) + \sum\limits_{i=1}^{l} \mathbb{E}_{t_i \sim \mathit{Noise}(t)} [ \log \sigma(-\mathbf{v}_{t_i}^T \mathbf{v}_{t_I}) ],$$ where $\sigma(x) = 1 / (1 + \exp(-x))$, $\mathit{Noise}(t)$ is the noise distribution of topics, and $l$ is the number of negative samples of each topic. Thus the task is to distinguish the target topic $t_O$ from draws from the noise distribution $\mathit{Noise}(t)$. We also do subsampling~\cite{mikolov2013distributed} of frequent topics in our model to counter the imbalance between rare and frequent topics: each topic $t_i$ in the training set is discarded with probability computed by the formula: $$P(t_i) = 1 - \sqrt{\frac{\delta}{F(t_i)}},$$ where $F(t_i)$ is the frequency of topic $t_i$ and $\delta$ is a chosen threshold, typically around $10^{-5}$.

\begin{algorithm}
\caption{Top-$k$ Area Topics Extraction}
\label{ate}
\KwIn{Area $r$, knowledge base $\mathit{KB}=(C,R,X)$, the number $k$ of topics to extract, general weights $\{w_j^r\}_j$ of topics in $C$.}
\KwOut{The top-$k$ topics set $T^k$.}
$T = \emptyset$;

\While{$|T|<k$} 
  {
      $m = -1$;
      
      \ForEach{$t_i \in C \setminus T$}
      {
        $s$ = 0;

        \ForEach{$ t_j \in C $}
        {
            $s$ += $w_j^r D(T\cup\{t_i\}, \{t_j\}, r)$;
        }
        
        \If{$s > m$}
        {
            $t = t_i$;
            
            $m = s$;
        }

      }
      $T = T \cup \{t\}$;
  }
return $T$\;
\end{algorithm}

\subsection{Top-$\boldsymbol{k}$ Area Topics Extraction} \label{subsec:topic_extraction}
As stated in section~\ref{sec:pf}, our problem is formulated as an optimization problem:
\begin{equation}
\label{optimal}
T^{k*}={\argmax_{T^k \in C,|T^k|=k}}{D(T^k, C, r)},
\end{equation}
where $D(T^k, C, r)$ is a function that denotes the degree of how well a set of $k$ topics $T^k \subseteq C$ can represent all topics in $C$ on the given area $r$.

\paragraph{NP-hardness.} We first prove the problem is NP-hard by reducing Dominating Set Problem\cite{karp1972reducibility,gary1979computers} to this problem as follows.
\begin{proof}
For $t_i, t_j \in C$, we first define the relativeness between $t_i, t_j$ as $I(t_i, t_j) \in [-1,1]$, and if $I(t_i, t_j) \ge 0$, we assign an undirected edge $e_{ij}$ between $t_i$ and $t_j$; otherwise, there is no edge between $t_i$ and $t_j$. Thus we get get an undirected graph $G=(C, E)$ of all concepts in $C$, where $E$ is the set of all edges in $G$.

Then we define $D(T^k, \{t_i\}, r)$ as: $D(T^k, \{t_i\}, r) = 1$ if $\exists t_a \in T^k$ such that $e_{ai} \in E$; $D(T^k, \{t_i\}, r) = 0$ otherwise. And then we define $D(T^k, C, r)$ as: $$D(T^k, C, r) = \sum\limits_{t_i \in C} D(T^k, \{t_i\}, r).$$

We then show that if we can find the maximum value $M = \max D(T^k, C, r)$, we can also decide that for the given number $k \in \mathbb{Z}^+$, whether there exists a dominating set $G^s = (C^s, E^s)$ where $C^s \subseteq C$ and $E^s \subseteq E$ such that $|G^s| \le k$. The reduction process is as follows: we compare $M$ with $|C|$ which is the number  of concepts in $C$, and according to our definition of $D(T^k, C, r)$ and $D(T^k, \{t_i\}, r)$, it must hold that $M \leq |C|$. If $M = |C|$, then $\forall t_i \in C$, $\exists t_a \in T^k$ such that $e_{ai} \in E$, which means there exists a dominating set $G^s$ such that $|G^s| \le k$; if $M < |C|$, then $\exists t_i \in C$, such that $\forall t_a \in T^k$, $e_{ai} \not \in E$, which means there does not exist a dominating set $G^s$ such that $|G^s| \le k$.
\end{proof}

\paragraph{Heuristic Algorithm.} Since the problem is NP-hard, we propose an approximate heuristic algorithm in our model to solve it, as outlined in Algorithm~\ref{ate}, and detailed as follows. The main idea is that we select topics one by one, and in the $i$-th step, we select topic $t_j^*$ such that $$t_j^* = \argmax_{t_j \in C, t_j \not\in T^{i}} D(T^i\cup\{t_j\}, C, r),$$ where $T^{i}$ is the selected topics set before the $i$-th step. To calculate $D(T^i\cup\{t_j\}, C, r)$, we introduce the \textbf{general weight} $w_i^r$ to measure the importance of topic $t_i \in C$ in the given area $r$. We call $w_i^r$ \textbf{general weight} because this value will be set by utilizing some domain knowledge and may probably be not very precise and can only measure the importance of $c_i$ in area $r$ to some general extent. We will demonstrate the calculation process of $w_i^r$ in the following part. Then we define $D(T^i\cup\{t_j\}, C, r)$ as: $$D(T^i\cup\{t_j\}, C, r) = \sum\limits_{t_h \in T^i\cup\{t_j\}} \sum\limits_{t_l \in C} w_l^r D(\{t_h\}, \{t_l\}, r),$$ and define $D(\{t_h\}, \{t_l\}, r)$ as: $$D(\{t_h\}, \{t_l\}, r) = S(t_h, t_l),$$ where $S(t_h, t_l)$ represents the relativeness between $t_h$ and $t_l$. 

After we get the embeddings of topics in section~\ref{subsec:topic_embedding}, we can calculate $S(t_h, t_l)$ as follows: $$S(t_h, t_l) = \frac{\mathbf{v}_{t_h}\mathbf{v}_{t_l}}{\norm{\mathbf{v}_{t_h}}\norm{\mathbf{v}_{t_l}}},$$ where $\mathbf{v}_{t_h}$ and $\mathbf{v}_{t_l}$ are the embeddings of $t_h$ and $t_l$ respectively.


\paragraph{General Weight Calculation.} To calculate the general weight $w_i^r$ of topic $t_i \in C$ in the given area $r$, we incorporate the domain knowledge from an external large-scale knowledge base into our model. This shares a similar idea as Distant Supervision~\cite{mintz2009distant}. We still use Wikipedia as our knowledge base here, and use category information of the given area $r$ as the domain knowledge to help calculate $w_i^r$. The idea behind the calculation of general weight is that topics in shallower depth of subcategories of $r$ are probably more important in area $r$. More specifically, we calculate $w_i^r$ in the following steps:

\begin{itemize}
    \item Find the category that $r$ represents in Wikipedia, which is also denoted as $r$.
    \item For the given area $r$, extract its all subcategories $\mathit{SC} = \{\{t_0\}, \{t_{11}, t_{12}, \cdots\}, \{t_{21}, t_{22}, \cdots\}, \cdots\}$ recursively from Wikipedia, where $t_0$ is the root category $r$ and $t_{mn}$ represents the $n$-th subcategory in depth $m$.
    \item Calculate the general weight $w_{i}^r$ of topic $t_i \in C$ as: $w_{i}^r = g(n)$, where $n$ is the depth of topic $t_i$ in $r$'s subcategories if $t_i \in \mathit{SC}$; otherwise $w_{i}^r=0$ (or equivalently set $n=\infty$ if we want to put all topics in $\mathit{SC}$). $g(n)$ is a monotonically decreasing function of $n$, and can be selected empirically.
\end{itemize}

\begin{algorithm}
\caption{Fast Top-$k$ Area Topics Extraction}
\label{ate2}
\KwIn{Area $r$, high-quality candidate topics set $C_{d_1}^r$, contributive topics set $C_{d_2}^r$, the number $k$ of topics to extract, general weights $\{w_j^r\}_j$ of topics in $C$.}
\KwOut{The top-$k$ topics set $T^k$.}
$T = \emptyset$;

\While{$|T|<k$} 
  {
      $m = -1$;
      
      \ForEach{$t_i \in C_{d_1}^r \setminus T$}
      {
        $s$ = 0;

        \ForEach{$ t_j \in C_{d_2}^r $}
        {
            $s$ += $w_j^r D(T\cup\{t_i\}, \{t_j\}, r)$;
        }
        
        \If{$s > m$}
        {
            $t = t_i$;
            
            $m = s$;
        }

      }
      $T = T \cup \{t\}$;
  }
return $T$\;
\end{algorithm}

\subsection{Algorithmic Analysis} \label{subsec:aa}
We argue that Algorithm~\ref{ate} has at least an $(1-1/e)$-approximate of the original NP-hard problem. We first prove that the goal function of the original optimization problem is \textbf{non-negative}, \textbf{monotonically non-decreasing}, and \textbf{submodular}, and then we use these properties to prove its error bound. By definition the goal function is non-negative and monotonically non-decreasing; thus we only show its submodularity as follows.

\begin{proof}
As stated before, the problem is formulated as follows: $$T^{k*}={\argmax_{T^k \in C,|T^k|=k}}{D(T^k, C, r)},$$ where $D(T^k, C, r)$ is the goal function which represents the degree of how well topics set $T^k$ can represent $C$ in the given area $r$. For a given topic $t_i \not\in T^k$, we first denote $a_1 = D(T^k\cup\{t_i\}, C, r) - D(T^k, C, r)$, which means the increment to the goal function by adding $t_i$ to $T^k$. Then we add a topic $t_j \neq t_i$ and $t_j \not\in T^k$ to $T^k$, and denote $a_2 = D(T^k\cup\{t_j, t_i\}, C, r) - D(T^k\cup\{t_j\}, C, r)$. By the attribute of $D(T^k, C, r)$ we assume in section~\ref{sec:pf}, we have $a_2 \le a_1$, which means the goal function is submodular.

\end{proof}

Since the goal function of our problem is monotonically increasing, nonnegative and submodular, the solution generated by Algorithm~\ref{ate} is at least $(1-1/e)$ of the optimal solution~\cite{nemhauser1978analysis,kempe2003maximizing}.

\begin{table*}[ht]
  \caption{Performances of different methods in our experiment. Because there are only 8 nodes in the sub-tree of NLP in ACM CCS dataset, we leave it empty here. SE here corresponds to Software and its engineering in ACM CCS dataset.}
  \label{tab:performance1}
  \begin{adjustbox}{width=1\textwidth}
  \begin{tabular}{c|c|c|c|c|c|c|c|c|c|c|c}
    \toprule
    \multirow{2}{4em}{Dataset}&Area&\multicolumn{2}{c|}{AI}&\multicolumn{2}{c|}{CV}&\multicolumn{2}{c|}{ML}&\multicolumn{2}{c|}{NLP}&\multicolumn{2}{c}{SE} \\
    \cmidrule{2-12}
    &Metric & Presion@15 & MAP& Presion@15 & MAP& Presion@15 & MAP& Presion@15 & MAP& Presion@15 & MAP\\
    
    \midrule
    \multirow{4}{4em}{ACM CCS}&TFIDF & 0.1333 & 0.0144 & 0.0000 & 0.0000 & \textbf{0.3333} & 0.1560 & - & - & 0.2666 & 0.1286\\
    &LDA & 0.2667 & 0.1696 & 0.0000 & 0.0000 & 0.2667 & 0.1020 & - & - & 0.2000 & 0.1032\\
    &TextRank & \textbf{0.4000} & \textbf{0.2556} & 0.0000 & 0.0000 & \textbf{0.3333} & 0.1308 & - & - & 0.3333 & 0.1830\\
    &\model-1 & \textbf{0.4000} & 0.1551 & 0.0667 & 0.0056 & \textbf{0.3333} & 0.1183 & - & - & 0.4667 & 0.4137\\
    &\model-2 & \textbf{0.4000} & 0.1797 & \textbf{0.1333} & \textbf{0.0231} & \textbf{0.3333} & \textbf{0.1896} & - & - & \textbf{0.5333} & \textbf{0.4994}\\

    \midrule
    \multirow{4}{4em}{Microsoft FoS}&TFIDF & 0.1333 & 0.0333 & 0.0667 & 0.0222 & 0.4667 & 0.2864 & 0.1333 & 0.0933& 0.0667 & 0.0167\\
    &LDA & 0.2667 & 0.2130 & 0.2667 & 0.0989 & 0.4000 & 0.2901 & 0.1333 & 0.0889 & 0.2000 & 0.0375\\
    &TextRank & 0.4000 & 0.2600 & 0.2667 & 0.1302 & 0.4667 & \textbf{0.3529} & 0.1333 & 0.1000 & 0.3333 & 0.1077\\
    &\model-1 & 0.4000 & \textbf{0.4606} & 0.2667 & 0.2056 & \textbf{0.5333} & 0.3417 & 0.2000 & 0.1444 & 0.4000 & 0.1775\\
    &\model-2 & \textbf{0.4667} & 0.2193 & \textbf{0.3333} & \textbf{0.2405} & \textbf{0.5333} & 0.3522 & \textbf{0.2000} & \textbf{0.1511} & \textbf{0.4667} & \textbf{0.2658}\\

    \midrule
    \multirow{4}{4em}{Domain Experts}&TFIDF & 0.1333 & 0.0194 & 0.2000 & 0.0556 & 0.6667 & 0.4360 & 0.3333 & 0.2321& 0.2000 & 0.0952\\
    &LDA & 0.3333 & 0.2130 & 0.4000 & 0.1838 & 0.6000 & 0.4979 & 0.4000 & 0.2706& 0.2667 & 0.1261\\
    &TextRank & 0.4667 & 0.3750 & 0.4000 & 0.2183 & 0.7333 & 0.5666 & 0.4000 & 0.2853 & 0.4000 & 0.2189\\
    &\model-1 & 0.6000 & 0.4204 & 0.4667 & 0.3029 & 0.6667 & 0.5202 & 0.5333 & 0.3831 & 0.5333 & 0.4492\\
    &\model-2 & \textbf{0.7333} & \textbf{0.6097} & \textbf{0.6000} & \textbf{0.4389} & \textbf{0.8000} & \textbf{0.6710} & \textbf{0.5333} & \textbf{0.4020} & \textbf{0.6000} & \textbf{0.5394}\\
  
  \bottomrule
  \end{tabular}
  \end{adjustbox}
\end{table*}

\subsection{Fast Implementation} \label{subsec:fi}
The time complexity of Algorithm~\ref{ate} is $O(k |C|^2)$, where $k$ is the number of topics needed to be extracted and $|C|$ is the number of elements in $C$. In practical use, $k \leq 100$, but $|C|$ may be (tens of) millions of order of magnitude (i.e., we extract 9,355,550 candidate topics from Wikipedia as mentioned above). Thus Algorithm~\ref{ate} still seems infeasible and may take unbearable time to return results (which is actually the case in our experiments). However, we observe the following two facts:

\begin{itemize}
  \item Most of the candidate topics in the whole set are not relevant to a given area. 
  \item When the general weight of a topic is little enough, this topic's contribution to the whole sum ($s$ in Algorithm~\ref{ate}) may be little enough too. 
\end{itemize}

From the above two observations, we think of the following two strategies which can greatly speed up our algorithm:

\begin{itemize}
  \item We only keep topics within a depth $d_1$ in the given area's category as high-quality candidate topics from the original set.
  \item Since the general weight function $g(n)$ of a topic is monotonically decreasing with the topic's depth $n$, thus we can choose a depth $d_2$ (with a well-defined $g(n)$) such that the contributions of all topics below this depth are small enough and can be discarded without calculation.
\end{itemize}

And this can lead to a much faster algorithm with time complexity $O(k|C_{d_1}^r||C_{d_2}^r|)$, as summarized in Algorithm~\ref{ate2}, where $C_{d_1}^r$ and $C_{d_2}^r$ represent high-quality candidate topics set and contributive topics set respectively, and in practical use we have $|C_{d_1}^r| \ll |C|, |C_{d_2}^r| \ll |C|$.

\section{Experimental Results} \label{sec:er}
We train our model on one of the largest public knowledge base (Wikipedia). As there are no standard datasets with ground truth and also it is difficult to create such a data set of ground truth, for evaluation purpose, we collect three real-world datasets and choose five representative areas in computer science: Artificial Intelligence (AI), Computer Vision (CV), Machine Learning (ML), Natural Language Processing (NLP), and Software Engineering (SE) to compare the performance of our model with several alternative methods. But our model is not restricted to these areas and can be applied to any other areas theoretically. The datasets and codes are publicly available, and a demo is ready\footnote{https://github.com/thuzhf/FastKATE}---Inputs are: (1) $r$: area name in the form of a topical phrase (words are connected by a underline, such as ``artificial\_intelligence''). (2) $k$: the number of topics needed to be extracted. Outputs are: (1) Extracted $k$ topics ranked by and accompanied with their scores ($S_{t_h,t_l}$ in Section~\ref{subsec:topic_extraction}). (2) Running time.

\subsection{Datasets} \label{subsec:es}
We download Wikipedia data from wikidump\footnote{https://dumps.wikimedia.org/enwiki/latest/} as our knowledge base $\mathit{KB}$, use its (preprocessed) titles of all articles and categories as $C$, use the text of all articles as $X$ and use its category structures as $R$. After we preprocess the titles as stated in section~\ref{subsec:topic_embedding}, we get 9,355,550 candidate topics in $C$. As stated in section~\ref{subsec:topic_embedding}, we use full text of Wikipedia to train topics embeddings and view each topic as a whole in Word2Vec model, and we use Gensim\footnote{https://radimrehurek.com/gensim/models/word2vec.html} to help implement our model. The parameter settings are as follows: vector size $=200$, window size $=10$, min count of each topic $=2$, threshold ($\delta$) for downsampling $=0.0001$, min sentence length $=2$, num workers $=64$; for other parameters, we use default settings in Gensim. The collected three real-world datasets for evaluation are detailed as follows.

\textbf{ACM CCS classification tree.} ACM CCS classification tree\footnote{http://www.acm.org/about/class/class/2012} is a poly-hierarchical ontology and contains 2,126 nodes in total. In this tree (actually a directed acyclic graph), each node can be viewed as a topic and each non-leaf node has several children nodes as its sub-topics. Although different nodes may have different number and different granularity of nodes in its sub-tree, it still provides us a guidance that what may be top topics in a given area.

\textbf{Microsoft Fields of Study (FoS).} Microsoft Fields of Study (FoS) from its Microsoft Academic Graph (MAG)\footnote{https://www.microsoft.com/en-us/research/project/microsoft-academic-graph/} is a directed acyclic graph where each node also represent a topic and it contains 49,038 nodes in total. Each node in the graph is accompanied with a ``level'' representing its depth/granularity in the graph. The network has 4 different levels in total. Each node has super-nodes of different levels as its super-topics and each super-topic is accompanied with a confidence value. The confidences of all super-nodes of the same level of one topic sum to $1$. This dataset can also provide us a guidance that what may be top topics in a given area.

\textbf{Domain Experts Annotated Dataset.} As there are actually no standard datasets/benchmarks which perfectly match our problem, we also let domain experts directly annotate top-$k$ ($=15$) topics in the five given areas without giving any single dataset for reference.

For the first two datasets, we let domain experts select top-$k$ topics based on each given area's sub-topics to match our problem better. Because there may be too much nodes in certain area's sub-topics, we instruct domain experts to first select a larger set of topics than needed and then do secondary screening from them. Since we need to do annotations in all three datasets, we set up the following criterions to reduce subjectivity in the annotation process and help domain experts reach an agreement:


\begin{itemize}
  \item Selected topics should be more significant than other topics in the given area.
  \item Selected topics should cover the whole given area as far as possible. This implies that they should not be too similar with each other in the given area, such as Artificial Neural Networks and Neural Networks should be viewed as the same topic in AI area.
\end{itemize}

After we get the results of each domain expert, we count the number of each selected topic and rank them by their counts, and choose the top-$k$ from them as the ground truth of the given area. We empirically set $k=15$ in our experiment.

\subsection{Evaluation Metrics}
To quantitatively evaluate the proposed model, we consider the following two metrics.

\paragraph{Presion@k.} Since the number of extracted results are set to the same $k$ for domain experts and machines, we use Presion@$k$ to measure the performance of different methods. Since the order is also important in the extracted results, we introduce another metric as follows.

\paragraph{Mean Average Precision (MAP).} For a single result (such as a ranked list in our experiments), AP is defined as follows: $$AP = \sum_{k=1}^n \frac{P(k)}{\mathit{min}(m, n)}$$ where $m$ is the number of all correct items (i.e., the length of human-annotated ranked list); n is the length of the machine extracted ranked list (which is the same as $m$ in our experiments); $P(k)$ equals 0 when the $k$-th item is incorrect or equals the precision of the first $k$ items in the ranked list. MAP is then calculated by averaging the APs over all results.

\renewcommand{\multirowsetup}{\centering}

\begin{table*}[ht]
  \caption{Extracted topics ($k=15$) in AI area using different methods. \textbf{Bold} ones represent relatively \textbf{correct} ones.}
  \label{tab:performance1_more}
  \begin{adjustbox}{width=1\textwidth}
  \begin{tabular}{c|c|c|c}
    \toprule
    TFIDF&LDA&TextRank&\model-2\\
    \midrule
    Social Medium & \textbf{Robotics} & \textbf{Robotics} & \textbf{Machine Learning}\\
    User Interface & Virtual Reality & \textbf{Machine Learning} & \textbf{Computational Linguistics}\\
    Virtual Reality & \textbf{Machine Learning} & Semantic Web & \textbf{Knowledge Representation}\\
    Classification System & \textbf{Semantic Web} & \textbf{Natural Language Processing} & Artificial Life\\
    Facial Expression & Speech Recognition & Image Processing & Ubiquitous Computing\\
    World Wide Web & Turing Test & Virtual Reality & \textbf{Computer Vision}\\
    Remote Sensing & Usability & Speech Recognition & \textbf{Natural Language Processing}\\
    \textbf{Machine Learning} & User Interface & \textbf{Artificial Neural Network} & \textbf{Multi Agent System}\\
    Graphical User Interface & \textbf{Natural Language Processing} & User Interface & \textbf{Robotics}\\
    Image Processing & Knowledge Base & Usability & \textbf{Expert System}\\
    Unmanned Aerial Vehicle & World Wide Web & Knowledge Base & \textbf{Logic Programming}\\
    \textbf{Computer Vision} & Image Processing & \textbf{Knowledge Representation} & \textbf{Deep Learning}\\
    Knowledge Base & Optical Character Recognition & World Wide Web & Fuzzy Logic\\
    Aerial Photography & Handwriting Recognition & \textbf{Logic Programming} & \textbf{Artificial Neural Network}\\
    Speech Recognition & \textbf{Artificial Neural Network} & Fuzzy Logic & Computational Mathematics\\
  \bottomrule
  \end{tabular}
  \end{adjustbox}
\end{table*}

\subsection{Comparison Methods}
For each given area $r$, we first extract all its subcategories within a depth of $d_1$ ($=3$), and use them as candidate topics $C_{d_1}^r$. We then extract all articles $X^r \subseteq X$ of these candidate topics from Wikipedia for LDA and TextRank methods here, where $X$ represents the corpus in Wikipedia as introduced in section~\ref{sec:pf}.
\begin{itemize}
  \item \textbf{Topic TF-IDF (TFIDF):} We calculate each candidate topic's tf-idf~\cite{jones1973index} value in the whole Wikipedia corpus (viewing each article as a document), and rank all candidate topics by these values.
  \item \textbf{LDA:} We train LDA~\cite{blei2003latent} model on all documents in $X^r$. For each candidate topic $t_i \in C_{d_1}^r$ (note that this is in the form of a topical phrase, and it is not the extracted topics in LDA which is actually multinomial distributions over words), we calculate its weight $w_i$ as follows: $$w_i = \sum\limits_{j=1}^{|X^r|} \sum\limits_{l=1}^{K_\theta} \theta_{jl} \phi_{lt_i},$$where $K_\theta$ is the number of topics extracted by the LDA model, $\theta_{il}$ is the probability of $l$-th topic of the LDA model in the $i$-th article, and $\phi_{lt_i}$ is the probability of $t_i$ in $l$-th topic of the LDA model. When training, we remove those documents with $<100$ words. We utilize Gensim\footnote{https://radimrehurek.com/gensim/models/ldamodel.html} to help implement this model, and we use all default parameters of it except we set $K_\theta=500$.
  \item \textbf{TextRank:} We run TextRank~\cite{mihalcea2004textrank} algorithm on each article in $X^r$, and for each candidate topic $t_i \in C_{d_1}^r$ (in the form of a topical phrase), we calculate its weight $w_i$ as follows: $$w_i = \sum\limits_{j=1}^{|X^r|} \mathit{weight}_{ij},$$ where $\mathit{weight}_{ij}$ is the weight generated by TextRank of $t_i$ in $j$-th article of $X^r$.

  \item \textbf{\model:} This is our model outlined in Algorithm~\ref{ate2}. Due to the unbearable running time of Algorithm~\ref{ate}, we think it is impractical and thus do not compare its result with others. We empirically select $g(n) = \exp(4-n)$. We select two different settings of $d_1,d_2$ to compare their performances and time costs: (1) $d_1=d_2=3$ (denoted as \textbf{\model-1}), (2) $d_1=3, d_2=1$ (denoted as \textbf{\model-2}).
\end{itemize}

\begin{table}[ht]
  \caption{Average running time of our two models over 100 times runs. We can see that \model-2 can return results in real-time.}
  \label{tab:running_time}
  \begin{adjustbox}{width=1\columnwidth}
  \begin{tabular}{c|c|c|c|c|c}
    \toprule
    Area&AI&CV&ML&NLP&SE\\
    \midrule
    \model-1 & 34.217s & 4.533s & 0.262s & 0.101s & 36.675s\\
    \model-2 & 0.593s & 0.122s & 0.025s & 0.009s & 0.093s\\
  \bottomrule
  \end{tabular}
  \end{adjustbox}
\end{table}

\subsection{Results and Analysis} \label{subsec:qr1}
\paragraph{Accuracy Performance.} Table~\ref{tab:performance1} lists the performances of different methods used in the problem of extracting top-$k$ topics in a given area. In terms of Precision@$k$, our model \model-2 performs consistently the best on all three datasets and in all five areas. In terms of MAP, our model \model-2 performs the best in $11/14$ cases. This suggests our model can not only extract more correct top-$k$ topics but also rank them in more accurate order. We can also see that \model-1 (it is different from \model-2 only in parameter settings) performs the second best in most cases, which suggests that even with different parameter settings, our model is still very effective comparing to other methods so that our model is also robust.

We note that average performances of all methods on the first two datasets (ACM CCS and Microsoft FoS) are worse than on the third dataset, which is annotated by domain experts specially for this problem. This is easy to understand since there are actually no existing datasets/benchmarks that can perfectly match this problem, and although the first two datasets are highly-related to the problem, they are not specialized for this purpose. And this is the reason that we annotate our own datasets with the help of domain experts directly, and we think the third dataset is more capable of reflecting the performances of different methods on this problem.

It is beyond our expectation that \model-2 performs better than \model-1 in most cases, because \model-2 uses smaller contributive topics set than \model-1 ($d_2=1<3$) and thus seems accessing less information than \model-1. We think one possible reason is that the contributive topics set becomes more noisy when they go deeper, and when we restrict the depth to only one, we have cleaner data and thus may get better results. Besides, as stated in section~\ref{subsec:fi}, when the depth is smaller, our algorithm runs faster. We record the average running time of our two models over 100 times runs on all five areas in Table~\ref{tab:running_time}. We can see that \model-2 is $>10\times$ faster than \model-1 and can return results in real-time.



\subsection{Case Study} \label{subsec:qr2}
Table~\ref{tab:performance1_more} lists extracted topics in AI area using TFIDF, LDA, TextRank and \model-2 respectively. We can see that most extracted topics by our model are of high-quality and are more convincing compared to other methods.


\section{Related Work} \label{sec:rw}
Our work is mainly related to the work from the following three aspects: topic modeling, automatic keyphrase extraction and word/phrase embedding. (1) \textbf{Topic Modeling}. Topic modeling has been widely used to extract topics from large-scale scientific literature~\cite{blei2003latent,griffiths2004finding,steyvers2007probabilistic}. Topics in these models are usually in the form of multinomial distributions over words, which makes it hard for researchers to identify which specific topics these distributions stand for~\cite{mei2007automatic}. To address this challenge, many work has been conducted to find an automatic or semi-automatic way to label these topic models~\cite{mei2007automatic,ramage2009labeled,lau2011automatic}, which alleviate this problem to some extent. (2) \textbf{Automatic Keyphrase Extraction.} There are mainly two approaches to extracting keyphrases: supervised and unsupervised. In supervised methods, the keyphrase extraction problem is usually re-casted as a classification problem~\cite{witten1999kea,turney2000learning} or as a ranking problem~\cite{jiang2009ranking}. Existing unsupervised approaches to keyphrase extraction can be categorized into four groups~\cite{hasan2014automatic}: graph-based ranking~\cite{mihalcea2004textrank}, topic-based clustering~\cite{liu2009clustering}, simultaneous learning~\cite{wan2007towards} and language modeling~\cite{tomokiyo2003language}. (3) \textbf{Word/Phrase Embedding.} Feature learning has been extensively studied by the machine learning community under various headings. In natural language processing (NLP) area, feature learning of words/phrases is usually referred to as word/phrase embedding, which means embedding words/phrases into a latent feature space~\cite{roweis2000nonlinear,mikolov2013distributed}. This method can help calculate relations/similarities between words/phrases. In our work, we embed topics into a latent feature space, which is similar to this line of work.

\section{Conclusion} \label{sec:c}
In this paper, we formally formulate the problem of top-$k$ area topics extraction. We propose \textbf{\model} in which topics have both explicit and latent representations. We leverage a large-scale knowledge base (Wikipedia) to learn topic embeddings and use this kind of representations to help capture the representativeness of topics for given areas. We develop a heuristic algorithm together with a fast implementation to efficiently solve the problem and prove it is at least $(1-1/e)$ of the optimal solution. Experiments on three real-world datasets and in five different areas validate our model's effectiveness, robustness, real-timeness (return results in $<1$s), and its superiority over other methods. In future, we plan to integrate more knowledge bases and also try to apply our model to a broader range of problems.


\bibliography{zhang}
\bibliographystyle{aaai}
\end{document}